\documentclass[preprint,12pt]{elsarticle}

\usepackage{amssymb}

\usepackage{amsmath}
\usepackage{bbm}
\usepackage{dsfont}

\usepackage{multirow}
\usepackage{verbatim}

\usepackage{floatrow}
\newfloatcommand{capbtabbox}{table}[][\FBwidth]

\usepackage{algorithmic}
\usepackage{algorithm}

\usepackage[pagebackref=true,breaklinks=true,colorlinks,bookmarks=false]{hyperref}

\journal{Image and Vision Computing}

\begin{document}

\begin{frontmatter}



\title{Spatial Likelihood Voting with Self-Knowledge Distillation for Weakly Supervised Object Detection}


\author[1]{Ze Chen}
\author[5]{Zhihang Fu}
\author[5]{Jianqiang Huang}
\author[5]{Mingyuan Tao}

\author[1]{Rongxin Jiang}
\author[1]{Xiang Tian}
\author[1]{Yaowu Chen\corref{cor1}}
\ead{cyw@mail.bme.zju.edu.cn}
\author[5]{Xian-Sheng Hua}

\cortext[cor1]{Corresponding author}

\affiliation[1]{organization={Zhejiang University, Institute of Advanced Digital Technology and Instrument},
            addressline={No.38 Zheda Road}, 
            city={Hangzhou},
            postcode={310027},
            country={China}}




\affiliation[5]{organization={Alibaba Group},
            city={Hangzhou},
            postcode={311121}, 
            country={China}}
        
\date{}

\begin{abstract}
Weakly supervised object detection (WSOD), which is an effective way to train an object detection model using only image-level annotations, has attracted considerable attention from researchers.
However, most of the existing methods, which are based on multiple instance learning (MIL), tend to localize instances to the discriminative parts of salient objects instead of the entire content of all objects.
In this paper, we propose a WSOD framework called the Spatial Likelihood Voting with Self-knowledge Distillation Network (SLV-SD Net).
In this framework, we introduce a spatial likelihood voting (SLV) module to converge region proposal localization without bounding box annotations.
Specifically, in every iteration during training, all the region proposals in a given image act as voters voting for the likelihood of each category in the spatial dimensions.
After dilating the alignment on the area with large likelihood values, the voting results are regularized as bounding boxes, which are then used for the final classification and localization.
Based on SLV, we further propose a self-knowledge distillation (SD) module to refine the feature representations of the given image.
The likelihood maps generated by the SLV module are used to supervise the feature learning of the backbone network, encouraging the network to attend to wider and more diverse areas of the image.
Extensive experiments on the PASCAL VOC 2007/2012 and MS-COCO datasets demonstrate the excellent performance of SLV-SD Net.
In addition, SLV-SD Net produces new state-of-the-art results on these benchmarks.
\end{abstract}




\end{frontmatter}


\section{Introduction}
\label{introduction}
Object detection is a fundamental and important task in computer vision.
Its aim is to locate all the objects of interest with tight bounding boxes in a given image and classify each one.
With the development of convolutional neural networks (CNNs)~\cite{simonyan2014very,he2016deep,krizhevsky2012imagenet,lecun1998gradient} and large-scale annotated datasets~\cite{everingham2015pascal,lin2014microsoft,russakovsky2015imagenet,shao2019objects365}, object detection performance has been substantially improved in the last decade~\cite{girshick2015fast,girshick2014rich,redmon2016you,lin2017feature,liu2016ssd,ren2015faster,tan2020efficientdet}.
However, it is time-consuming and labor-intensive to use a fully supervised learning mechanism to annotate accurate object bounding boxes for a large-scale dataset.
Therefore, weakly supervised object detection (WSOD), which only use image-level annotations for training, is a promising practical solution and has attracted the attention of researchers in recent years.

\begin{figure}
\begin{center}
\includegraphics[width=0.8\linewidth]{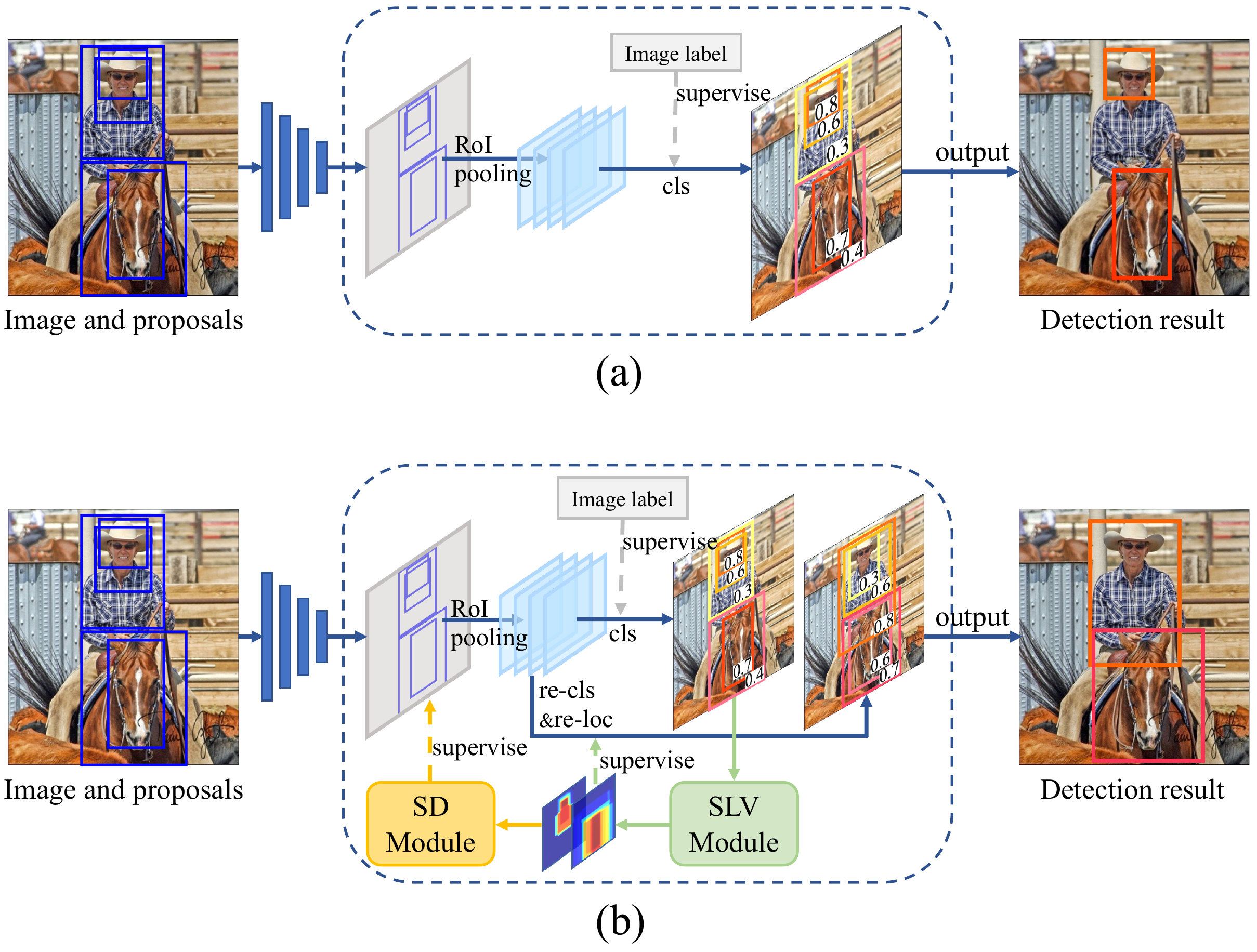}
\end{center}
   \caption{
    {Comparison between a common MIL-based method and SLV-SD Net.
   (a) A common MIL-based method is easy to localize instances to their discriminative parts instead of the whole content.
   (b) SLV-SD Net with SLV module and SD module can detect accurate bounding boxes of objects.}
    }
\label{fig:fig-1}
\end{figure}

The multiple instance learning (MIL) paradigm is followed in most recent WSOD studies~\cite{bilen2016weakly,cinbis2016weakly,ren2015weakly,tang2018pcl,tang2017multiple,wei2018ts2c}.
Each image is treated as a bag, and its region proposals are treated as instances.
An instance classifier is trained to distinguish at least one instance from the rest for each positive category.
However, these studies only focus on feature representations for instance classification and do not consider the localization accuracy of the predictions.
Furthermore, owing to the constraints of MIL, less salient objects are easily ignored when there are multiple objects of the same category in an image.
Moreover, the most discriminative parts of an object are always detected instead of the whole content, as shown in Fig.~\ref{fig:fig-1} (a).
Consequently, there is still a large performance gap between weakly supervised and fully supervised methods.

In this paper, we propose spatial likelihood voting with self-knowledge distillation network (SLV-SD Net) to accurately detect objects using a weakly supervised learning mechanism.
The spatial likelihood voting (SLV) module converges the region proposal localization without the need for bounding box annotations.
Using the SLV module, the self-knowledge distillation (SD) stage refines the feature representations of the training images to circumvent the above-mentioned problems of MIL, as shown in Fig.~\ref{fig:fig-1} (b).

Specifically, the SLV operation consists of instance selection, spatial probability accumulation, and high likelihood region voting.
Unlike previous methods, which fix the position of their region proposals, the region proposals in SLV act as voters in each iteration during training, voting for the likelihood of each category in the spatial dimensions.
Then the voting results, which will be used for the re-classification and re-localization, are regularized as bounding boxes by dilating the alignment on the areas with large likelihood values.
An adaptive search algorithm is proposed to obtain accurate bounding boxes from the likelihood maps of each image.
Through the generation of the voting results, the proposed SLV reformulates the instance classification problem as a multi-tasking problem.
The SLV enables WSOD methods to learn classification and localization simultaneously.
The classification and localization tasks facilitate each other, which leads to better localization and classification results and reduces the performance gap between weakly supervised and fully supervised object detection.
Additionally, from the perspective of refining feature distribution, we propose an SD stage based on the SLV module.
The likelihood maps of each image are used to supervise the feature learning of the backbone network.
The SD stage explicitly encourages the backbone network to focus on the less salient objects and less discriminative parts of an object, which better handles the ill-posed nature of WSOD.

Extensive experiments were conducted on the PASCAL VOC~\cite{everingham2015pascal} and MS-COCO~\cite{lin2014microsoft} datasets to evaluate the effectiveness of our method.
The proposed SLV-SD Net produces new state-of-the-art results in these benchmarks.

The contributions of this paper are summarized as follows:
\begin{enumerate}
\item We propose an SLV module to converge the region proposal localization with only image-level annotations.
The proposed SLV evolves the instance classification problem into a multi-tasking problem.
Moreover, an adaptive search algorithm is proposed to obtain accurate voting results. 
\item We propose an SD stage to supervise the feature learning of the backbone network. 
This helps handle the ill-posed nature of WDOD and enhances the performance of the final detection model.
\item Extensive experiments were conducted on different datasets to evaluate the method. 
The results suggest that sophisticated localization fine-tuning and feature refinement should be a promising direction for future exploration.
\end{enumerate}

A preliminary version of this work was published previously~\cite{chen2020slv}.
The present work adds to the initial version in a number of ways.
First, we improve the SLV module in the previous manuscript by introducing an adaptive search algorithm to obtain accurate voting results for each training image.
Second, we extend the previous method with a SD stage.
This stage, which encourages the network to pay attention to more diverse and wider areas for better detection performance, is designed to cope with the ill-posed nature of WSOD.
Third, a more comprehensive experimental analysis is presented in Section~\ref{experiment}.
A comparison with a number of recent WSOD methods on different benchmarks and different backbones demonstrates that our method is state of the art.

\section{Related Work}
\subsection{Weakly Supervised Object Detection}
MIL is a classical weakly supervised learning problem that is currently the main approach to WSOD.
MIL treats each training image as a ``bag'' and candidate proposals as ``instances.''
The objective of MIL is to train an instance classifier to select positive instances from this bag.
With the development of convolutional neural networks (CNNs), many methods~\cite{bilen2016weakly,diba2017weakly,kantorov2016contextlocnet,tang2017deep} combine CNN and MIL to solve the WSOD problem.
For example, Bilen and Vedaldi~\cite{bilen2016weakly} proposed a representative two-stream weakly supervised deep detection network (WSDDN), which can be trained with image-level annotations in an end-to-end manner. 
Based on the architecture in~\cite{bilen2016weakly},~\cite{kantorov2016contextlocnet} exploited the contextual information from regions around the object as a supervisory guidance for WSOD.

In practice, it has been found that MIL solutions often converge to discriminative parts of the objects. 
This is because the loss function of MIL is non-convex and thus MIL solutions usually become stuck in local minima.
To address this problem, Tang \textit{et al.}~\cite{tang2017multiple} combined WSDDN with a multi-stage classifier and proposed an OICR algorithm to help their network pay attention to larger regions of the objects during training.
Moreover, building on~\cite{tang2017multiple}, Tang \textit{et al.}~\cite{tang2018pcl} subsequently introduced proposal cluster learning (PCL) and used the proposal clusters for supervision that indicates the rough locations where objects are most likely to appear.
In~\cite{wan2018min}, the aim of Wan \textit{et al.} was to reduce the randomness of localization during learning.
In~\cite{zhang2018zigzag}, Zhang \textit{et al.} added curriculum learning using the MIL framework.
From the perspective of optimization, Wan \textit{et al.}~\cite{wan2019c} introduced the continuation method and smoothed the loss function of MIL to alleviate the problems caused by non-convexity.
In~\cite{gao2019utilizing}, Gao \textit{et al.} employed the instability of MIL-based detectors and designed a multi-branch network with orthogonal initialization.

In addition, there have been many attempts~\cite{gao2018c,arun2019dissimilarity,li2019weakly,yang2019towards,zhang2018w2f,tang2018weakly,ren2020instance,huang2020comprehensive} to improve the localization accuracy of weakly supervised detectors from other perspectives.
Gao \textit{et al.}~\cite{gao2018c} proposed using a per-class object count to identify correct high-scoring object bounding boxes, and then used these high-quality regions to refine a weakly supervised object detector.
Arun \textit{et al.}~\cite{arun2019dissimilarity} substantially improved performance by employing a probabilistic objective to model the uncertainty in the location of objects.
In~\cite{li2019weakly}, Li \textit{et al.} proposed a segmentation-detection collaborative network that uses segmentation maps as prior information to supervise the learning of object detection.
In a similar manner,~\cite{shen2019cyclic} presented a cyclic guidance learning method for segmentation and detection.
In~\cite{zhang2018w2f}, Zhang \textit{et al.} mined accurate pseudo ground-truth from a well-trained MIL-based network to train a fully supervised object detector.
In contrast, the work of Yang \textit{et al.}~\cite{yang2019towards} integrates WSOD and Fast-RCNN re-training into a single network that can jointly optimize regression and classification.
In~\cite{yin2021instance}, Yin \textit{et al.} proposed an instance mining framework with class feature bank, which is helpful to improve the region proposal selection of the MIL branch.
In~\cite{tang2018weakly}, Tang \textit{et al.} proposed a region proposal network in weakly supervised mechanism to improve the quality of proposals.
In~\cite{huang2020comprehensive}, Huang \textit{et al.} proposed a sophisticated framework which combines proposal features from different transformations of the same image and different network layers to activate the whole body of the object in this proposal.

\subsection{Knowledge Distillation}
In~\cite{bucilua2006model}, Buciluǎ \textit{et al.} first proposed transferring information from a large model to smaller, faster models without a significant loss in accuracy.
This learning paradigm was popularized by Hinton \textit{et al.} in~\cite{hinton2015distilling} as ``knowledge distillation.''
Hinton \textit{et al.} proposed supervising a small student model using the soft targets output by a large teacher model.
This student model could achieve performance that was competitive with or even better than the teacher model.
In addition to transferring the output of the last layer, the intermediate representations could also be transferred.
In~\cite{romero2014fitnets}, Romero \textit{et al.} proposed using the intermediate representations from the hidden layer of the teacher model to guide the training of the student model.
In~\cite{zagoruyko2016paying}, Zagoruyko and Komodakis proposed properly defining the attention of CNNs and transferring the attention maps from a powerful teacher model to student models.

In addition to transferring the knowledge from one model to another, the knowledge could also be transferred within the same model, in a process called self-distillation.
In~\cite{zhang2019your}, Zhang \textit{et al.} proposed a self-distillation framework that transfers knowledge from the deeper portion of the network to the shallow portion.
Similarly, Hou~\cite{hou2019learning} proposed a self-attention distillation method for lane detection.
In~\cite{yang2019snapshot}, Yang \textit{et al.} proposed an interesting self-distillation framework that transfers knowledge from the earlier epochs to later epochs. 

\begin{figure*}[ht]
\begin{center}
\includegraphics[width=1.0\linewidth]{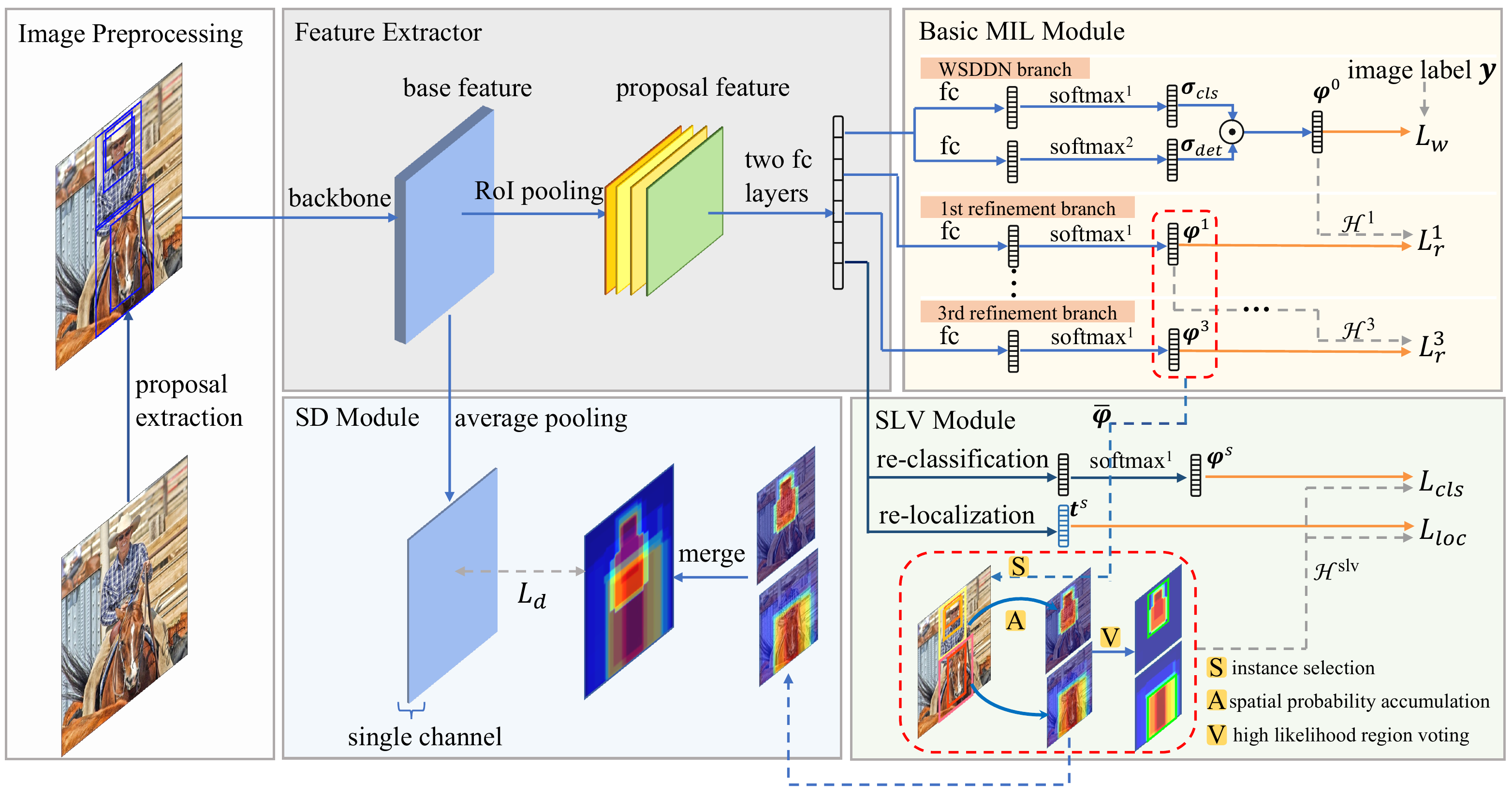}
\end{center}
   \caption{Network architecture of our method. 
    A CNN backbone with region-of-interest (RoI) pooling is used to extract the feature of each proposal.
    Next, the proposal features pass through two fully connected layers and the generated feature vectors are branched into the Basic MIL module and SLV module.
    In the Basic MIL module, there are three refinement branches and one WSDDN branch.
    The average classification scores of the three refinement branches are fed into the SLV module to generate supervisions.
    Here, $softmax^1$ is the softmax operation over the categories and $softmax^2$ is the softmax operation over the proposals.
    In the SD module, the likelihood maps of positive categories are merged into one map in order to supervise the feature learning of the backbone network.
    }
\label{fig:architecture}
\end{figure*}

\section{Method}
The proposed SLV-SD Net is introduced in detail in this section and its architecture is shown in Fig.~\ref{fig:architecture}.
Overall, the framework consists of three parts:
1) An MIL-based network that extracts the image feature maps and produces proposal score matrices; 
2) An SLV module that utilizes the proposal score matrices to generate supervisions for the re-classification and re-localization branches;
3) An SD module that refines the feature distribution of the training image based on the SLV module.

\subsection{Basic MIL Module} \label{section3-1}
With image-level annotations, many existing methods~\cite{bilen2015weakly,bilen2016weakly,cinbis2016weakly,kantorov2016contextlocnet} detect objects based on an MIL network. 
In this work, we follow the method in~\cite{bilen2016weakly}, which proposed a two-stream WSDDN to train the instance classifier.
For an input image and its region proposals, a CNN backbone with region-of-interest (RoI) pooling extracts the proposal features.
Then, the proposal features are branched into two streams, namely the classification branch and detection branch.
For the classification branch, the region proposal features are passed through a fully connected (fc) layer to produce score matrix ${\boldsymbol{X}}^{cls}$ $\in$ $\mathbb{R}^{C\times R}$, where $C$ denotes the number of image categories and $R$ denotes the number of proposals.
Next, a softmax operation over categories is performed to produce $\boldsymbol{\sigma}_{cls} ( {\boldsymbol{X}}^{cls} )$, $[\boldsymbol{\sigma}_{cls} ( {\boldsymbol {X}}^{cls} )]_{c,r} = \frac{e^{{\boldsymbol{X}_{c,r}^{cls}}}} { \sum_{k=1}^{C}e^{\boldsymbol{X}_{k,r}^{cls}}}$.
Similarly, score matrix ${\boldsymbol{X}}^{det}$ $\in$ $\mathbb{R}^{C\times R}$ is produced by another fc layer for the detection branch, but $\boldsymbol{\sigma}_{det}({\boldsymbol {X}}^{det})$ is generated through a softmax operation over proposals rather than categories:
$[ \boldsymbol{\sigma}_{det}( {\boldsymbol {X}}^{det} ) ]_{c,r} = \frac{e^{{\boldsymbol{X}_{c,r}^{det}}}} { \sum_{k=1}^{ R }e^{\boldsymbol{X}_{c,k}^{det}}}$.
The score of each proposal is computed by the element-wise product:
$\boldsymbol{\varphi}^0 = \boldsymbol{\sigma}_{cls} ( {\boldsymbol {X}}^{cls} )\odot \boldsymbol{\sigma}_{det} ( {\boldsymbol {X}}^{det} ) $.
Finally, the classification score on category $c$ of the input image is computed through the summation over all proposals:
$\phi_c = \sum_{r=1}^{R}\boldsymbol{\varphi}_{c,r}^0$.
The label of a training image is denoted as ${\boldsymbol {y} } = \left [ y_1,y_2,...,y_C \right ]^{T}$,
where $y_c = 1$ or 0 respectively indicates an image with or without objects of category $c$.
To train the instance classifier, the loss function is as follows.
\begin{equation}
L_w = -\sum_{c=1}^{C}\left \{ y_c\log \phi_c + (1 - y_c)\log (1 - \phi_c) \right \} \label{loss_w}
\end{equation}

Following the general paradigm of~\cite{tang2017multiple,tang2018pcl,yang2019towards}, we adopt PCL~\cite{tang2018pcl}, which embeds three instance classifier refinement branches, to obtain better instance classifiers.
As shown in Fig.~\ref{fig:architecture}, each refinement branch is supervised by the previous branch.
The output of the $k$-th refinement branch is $\boldsymbol {\varphi}^k \in \mathbb{R}^{(C + 1)\times R}$, where $(C + 1)$ is the number of classes $C$ plus the background.

Specifically, based on the output score $\boldsymbol {\varphi}^k$, several cluster centers are first selected.
Next, the remaining proposals are assigned to one of the cluster centers according to their intersection over union ($IoU$) values.
After that, all proposals are divided into several clusters (one for the background and the others for different instances).
Proposals in the same cluster (except for those in the background cluster) are spatially adjacent and associated with the same object.
PCL transforms the image-level annotations into cluster-level annotations, which are more informative.
With supervision $\mathcal{H}^k = \left \{ y_n^k \right \}_{n=1}^{N^k+1}$ ($y_n^k$ is the label of the $n$-th cluster), the refinement branch treats each cluster as a small bag.
Each bag in the $k$-th refinement branch is optimized by the following weighted cross-entropy loss.

\begin{equation}
\begin{aligned}
L^k = -\frac{1}{R}&( \sum_{n=1}^{N^k} s_n^k M_n^k \log \frac{\sum\limits_{r \in \mathcal{C}_n^k} \boldsymbol{\varphi}_{y_n^k,r}^k }{M_n^k} \\
&+ \sum\limits_{r \in \mathcal{C}_{N^k+1}^k} \lambda_r^k \log \boldsymbol{\varphi}_{C+1,r}^k) \label{loss_k}
\end{aligned}
\end{equation}
where $s_n^k$ and $M_n^k$ are respectively the confidence score of the $n$-th cluster and the number of proposals in the $n$-th cluster, and $\boldsymbol{\varphi}_{c,r}^k$ is the predicted score of the $r$-th proposal.
Here, $r \in \mathcal{C}_n^k$ indicates that the $r$-th proposal belongs to the $n$-th proposal cluster, $\mathcal{C}_{N^k+1}^k$ is the cluster for the background, and $\lambda_r^k$ is the loss weight, which is the same as the confidence of the $r$-th proposal.

\subsection{Spatial Likelihood Voting} \label{section 3-2}
For MIL-based methods, it is difficult to determine the most appropriate bounding boxes from all the proposals for an object.
The proposal that obtains the highest classification score often covers a discriminative part of an object, whereas many of the proposals that cover the larger parts of the object tend to have lower scores.
Therefore, choosing the proposal with the highest score as the detection result within the MIL framework is not a satisfactory solution.
We observe that in the overall distribution, high-scoring proposals always cover at least part of an object.
Consequently, we use the spatial likelihood of all proposals, which implies the boundaries and categories of the objects in an image.
This subsection presents the SLV module, which performs both classification and localization refinement instead of only instance classification.

The SLV module can be conveniently integrated into any proposal-based detector and can be optimized jointly with the fundamental detector.
The concept of SLV is to establish a bridge between the classification task and localization task by coupling together the spatial information and category information of all proposals.
During training, the SLV module takes the classification scores of all proposals and calculates their spatial likelihood to generate supervision $\mathcal{H}^{slv}\left ( \bar{\boldsymbol{\varphi}}, \boldsymbol{{y}}\right )$, where $\bar{\boldsymbol{\varphi}} = \left ( \sum_{k=1}^{3} \boldsymbol{\varphi}^k \right )/3$.
{
$\mathcal{H}^{slv}\left ( \bar{\boldsymbol{\varphi}}, \boldsymbol{{y}}\right )$ will be used to supervise the training of re-classification and re-localization branches.
We use the notation $\mathcal{H}^{slv}$ for simplification, dropping the dependence on $\bar{\boldsymbol{\varphi}}$ and $\boldsymbol{{y}}$.}

\begin{algorithm}[tb] 
\caption{Bounding boxes generation.} 
\label{alg:algo1} 
\begin{algorithmic}[1] 
\REQUIRE ~~\\ 
Likelihood map $\boldsymbol{{M}}^c$.
\ENSURE ~~\\ 
Bounding boxes $\mathcal{G}_c$.
\STATE Initialize $\mathcal{G}_c = \varnothing$.\\
Initialize step size $s = 5$.\\
Initialize $\varepsilon = 0.05$.\\
Initialize $p_{min} = 1$.\\

\STATE $T_{search}$ = $\frac{1}{N}\sum \mathds{1}\left \{\boldsymbol{M}^c_{i, j} > 0 \right \} \boldsymbol{M}^c_{i, j}$, $N$ is the number of non-zero elements in $\boldsymbol{M}^c$.
\WHILE{$max(\boldsymbol{{M}}^c)$ $>$ $T_{search}$}
\STATE Find the coordinates $(x_i, y_i)$ of the maximum element $max(\boldsymbol{{M}}^c)$.
\STATE $x_l = x_i$.
\WHILE{$T_{search} \leq \boldsymbol{{M}}^c_{y_i, x_l} \leq p_{min} + \varepsilon$}
\STATE $p_{min} = \boldsymbol{{M}}^c_{y_i, x_l}$.
\STATE $x_l = x_l - s$.
\ENDWHILE
\STATE Repeat operations similar to those of steps 5--9 to obtain $y_t, x_r, y_b$. 
\STATE Bounding box $b_t = (x_l, y_t, x_r, y_b)$.
\STATE $\mathcal{G}_c$ = $\mathcal{G}_c \cup b_t$.
\STATE $\boldsymbol{{M}}^c_{i, j} = 0$, where $y_t \leq i \leq y_b$ and $x_l \leq j \leq x_r$.
\ENDWHILE
\RETURN $\mathcal{G}_{c}$. 
\end{algorithmic}
\end{algorithm}

Formally, for an input image $\boldsymbol{\rm{I}}$ with label $\boldsymbol{y}$, generating $\mathcal{H}^{slv}_c$ when $y_c = 1$ consists of three steps.
In the first step, low-scoring proposals are filtered out to reduce the computation cost for training because they do not substantially affect the results.
The retained proposals are considered to surround the instances of category $c$ and are placed into set $\mathcal{B}_c$, $\mathcal{B}_c = \{b_{r}\ |\ \bar{\boldsymbol{\varphi}}_{c,r} > T_{score}\}$.

In the second step, we implement a spatial probability accumulator according to the predicted classification scores and locations of the proposals in $\mathcal{B}_c$.
In detail, a score matrix $\boldsymbol{{M}}^c$ $\in$ $\mathbb{R}^{H\times W}$ is constructed, where $H$ and $W$ are the height and width of the training image $\boldsymbol{\rm{I}}$, respectively.
All elements in $\boldsymbol{{M}}^c$ are initialized with zero.
Then, for each proposal $b_r \in \mathcal{B}_c$, the predicted score of $b_r$ is accumulated in $\boldsymbol{{M}}^c$ spatially.

\begin{equation}
\boldsymbol{{M}}^c_{i, j}=\sum_{\substack{r\;s.t.\;b_r \in \mathcal{B}_c, \\ \left (i,j\right) \in b_r }} \bar{\boldsymbol{\varphi}}_{c,r} \label{Mc}
\end{equation}
where $\left(i,j\right) \in b_r$ indicates pixel $\left (i,j\right)$ in proposal $b_r$.
{
The left term $\boldsymbol{{M}}^c_{i, j}$ is the element of $\boldsymbol{{M}}^c$ at coordinates $(i, j)$.
In the right term, $\bar{\boldsymbol{\varphi}}_{c,r}$ is the predicted score of the $r$-th proposal for class $c$.
Each element of $\boldsymbol{{M}}^c$ is the sum of predicted scores of those eligible proposals.
}
For proposals in $\mathcal{B}_c$, we calculate their likelihood in the spatial dimensions and the final value of the elements in $\boldsymbol{{M}}^c$ indicates the probability that an instance of category $c$ appears in that position.

\begin{algorithm}[tb] 
\caption{$\mathcal{H}^{slv}$ generation} 
\label{alg:algo2} 
\begin{algorithmic}[1] 
\REQUIRE ~~\\ 
Proposal boxes $\mathcal{B} = \left \{b_1,...,b_R\right \}$; proposal average scores $\bar{\boldsymbol{\varphi}}$;
image label vector $\boldsymbol{{y}} = \left [y_1,...,y_C\right ]^T$; image size $\left \{H, W\right \}$.
\ENSURE ~~\\ 
Supervision $\mathcal{H}^{slv}$.
\STATE Initialize $\mathcal{H}^{slv} = \varnothing$.\\
\FOR{$c = 1$ to $C$}
\STATE Initialize $\boldsymbol{{M}}^c$ with zero elements.
\IF{$y_c = 1$}
\STATE Initialize $\mathcal{B}_c = \varnothing$.\\

\FOR{$r = 1$ to $R$}
\IF{$\bar{{\boldsymbol{\varphi}}}_{c,r} > T_{score}$}
\STATE $\mathcal{B}_c = \mathcal{B}_c \cup b_r$.
\ENDIF
\ENDFOR
\STATE Construct $\boldsymbol{M}^c$ by Eq. (\ref{Mc}), see Section \ref{section 3-2}.
\STATE Scale the range of elements in $\boldsymbol{{M}}^c$ to $\left [ 0, 1 \right ]$.
\STATE Obtain bounding boxes $\mathcal{G}_c$ by Algorithm~\ref{alg:algo1}.
\STATE $\mathcal{H}_c^{slv} = \left \{ \mathcal{G}_c, c \right \}$.
\STATE $\mathcal{H}^{slv} = \mathcal{H}^{slv} \cup \mathcal{H}_c^{slv}$.
\ENDIF
\ENDFOR
\RETURN $\mathcal{H}^{slv}$. 
\end{algorithmic}
\end{algorithm}

Finally, the range of elements in $\boldsymbol{{M}}^c$ is scaled to $[0, 1]$, and an adaptive search algorithm is proposed to obtain accurate bounding boxes $\mathcal{G}_c$ from $\boldsymbol{{M}}^c$.
We find the most discriminative point in $\boldsymbol{{M}}^c$ first, which is the initial bounding box, and then expand this box in four directions (left, right, up, and down) until the stop conditions are met.
The details of this algorithm are summarized in Algorithm~\ref{alg:algo1}.
In contrast to the thresholding method in~\cite{chen2020slv}, the proposed adaptive search algorithm does not group multiple adjacent objects into one bounding box nor miss less salient objects.

\begin{figure}[t]
\begin{center}
\includegraphics[width=1.0\linewidth]{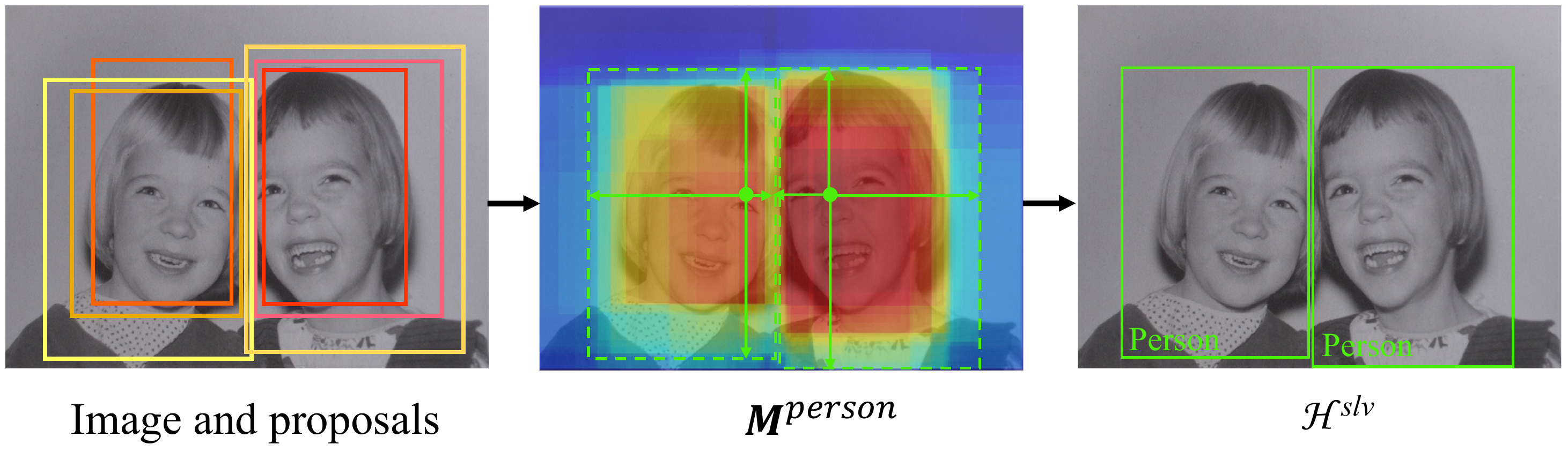}
\end{center}
\caption{A visualization example of SLV. The image label is $\left \{person\right \}$.
   The likelihood map $\boldsymbol{M}^c$ and $\mathcal{H}^{slv}$ are generated subsequently.}
\label{fig:slv present}
\end{figure}

The overall procedure for generating $\mathcal{H}^{slv}$ is summarized in Algorithm~\ref{alg:algo2}, and a visualization of SLV is shown in Fig.~\ref{fig:slv present}.
Supervision $\mathcal{H}^{slv}$ is an instance-level annotation, and we use multi-task loss $L_{s}$ on each labeled proposal to perform classification and localization refinement simultaneously.
The output of the re-classification branch is $\boldsymbol{\varphi}^{s} \in \mathbb{R}^{(C + 1)\times R}$ and the output of the re-localization branch is $\boldsymbol{t}^{s} \in \mathbb{R}^{[4\times (C+1)]\times R}$.
The loss of the SLV module is shown in Eq.~(\ref{slv_loss}), where $L_{cls}$ is the cross-entropy loss and $L_{loc}$ is the smooth L1 loss.
\begin{equation}
L_{s} = L_{cls}(\boldsymbol{\varphi}^{s}, \mathcal{H}^{slv}) + L_{loc}(\boldsymbol{t}^{s}, \mathcal{H}^{slv}) \label{slv_loss}
\end{equation}

\subsection{Self-Knowledge Distillation} \label{section 3-3}
In MIL, the model needs to find at least one positive instance from the bag for a positive category.
However, this constraint encourages the model to focus on the most salient object and ignore the less salient objects when multiple objects with the same category exist in the image.
Meanwhile, because of the biased distribution of training instances, the model is prone to converging to the most discriminative parts of an object rather than to the whole content.
The above two problems cause the feature representations of an image to be concentrated in the most discriminative parts of the most salient objects.
To address these problems, in this subsection, we introduce an SD module based on the SLV.

The likelihood maps generated in Section~\ref{section 3-2} are used to supervise the training of the backbone network.
As illustrated in Fig.~\ref{fig:architecture}, the image feature map extracted by the backbone network is $\boldsymbol{F} \in \mathbb{R}^{h\times w\times d}$, where $d$ is the depth of the feature map.
This feature map $\boldsymbol{F}$ is transformed into a single channel map $\boldsymbol{F}^{'} \in \mathbb{R}^{h\times w}$ by channel-wise average pooling as follows.
\begin{equation}
\boldsymbol{F}^{'}_{i,j} = \frac{1}{d}\sum_{k=1}^{d}(\boldsymbol{F}_{i,j,k}) \label{avg_pool}
\end{equation}
We then merge the likelihood maps into one map and resize it to the same spatial size as $\boldsymbol{F}^{'}$ as follows.
\begin{equation}
\boldsymbol{M} = R(\frac{1}{\left\| \boldsymbol y \right \|_0}\sum_{c=1}^{C}\mathds{1}\left \{y_c = 1 \right \}\boldsymbol{M}^{c}) \label{merge map}
\end{equation}
Here, $\boldsymbol y$ is the image label and $\left\| \boldsymbol y \right \|_0$ is the number of non-zero elements in $\boldsymbol y$.
Further, $R(\cdot)$ is the resize operation and $\boldsymbol{M} \in \mathbb{R}^{h\times w}$.
The SD loss is defined as follows.
\begin{equation}
L_d = \left \| \boldsymbol{M} - \boldsymbol{F}^{'} \right \|_2 \label{loss_d}
\end{equation}

Finally, the overall loss function of our SLV-SD Net is expressed as follows.
\begin{equation}
L = L_w + \sum_{k=1}^{3}L_r^k + w_sL_s + L_d \label{overall loss}
\end{equation}
Here, $w_s= S((i_{c} - i_{t}/2)/1000)$ is the loss weight for $L_s$, where $S(\cdot)$ is the sigmoid function.
Moreover, $i_{c}$ is the current number of iterations, and $i_{t}$ is the total number of training iterations.
Therefore, $w_s$ is initialized to zero and increases iteratively.
In the early stage of the training process, the basic MIL module is unstable, and the supervisions generated by the SLV module are noisy.
A small $w_s$ prevents the optimization of the basic MIL module from being affected by the SLV module.
As the training progresses, the basic MIL module will begin to classify the proposals better, and thus we can obtain stable classification scores for generating more precise supervisions.
We tried to adjust the weight between different losses and found that this strategy brings negligible performance improvement.
Therefore, we kept the weight of each component of the overall loss unchanged.

\section{Experiments and Analysis}\label{experiment}

\subsection{Datasets and Evaluation Metrics}
{We evaluated the proposed method, namely SLV-SD Net on three datasets, namely PASCAL VOC 2007, PASCAL VOC 2012~\cite{everingham2015pascal}, and MS-COCO~\cite{lin2014microsoft} following previous works.}
The PASCAL VOC 2007 dataset consists of 9,962 images, which are split into two sets: 5,011 for the training/validation ($trainval$) set and 4,951 for the $test$ set.
The PASCAL VOC 2012 dataset consists of 22,531 images, which are also split into two sets: 11,540 for the $trainval$ set and 10,991 for the $test$ set.
Both datasets contain 20 object categories.
For each dataset, the $trainval$ set is used for training and the $test$ set is used for testing.
{For MS-COCO dataset, there are 82,783 images for training and 40,504 images for testing.}
Only image-level annotations are available in the training process.

{
In the evaluation, mAP$_{0.5}$ (IoU threshold at 0.5) is used to evaluate the detection performance of our SLV-SD Net on the PASCAL VOC 2007 and 2012 $test$ sets.
The Correct Localization (CorLoc) metric is used to evaluate the localization accuracy of the proposed SLV-SD Net on the PASCAL VOC 2007 and 2012 $trainval$ sets.
The mAP$_{0.5}$ and mAP (averaged over IoU thresholds in $[.5:0.05:.95]$) is used to evaluate the performance of SLV-SD Net on MS-COCO dataset.
}

\subsection{Implementation Details}
The proposed SLV-SD Net is implemented based on a CNN model.
The selective search~\cite{uijlings2013selective} algorithm was adopted to generate about 2,000 proposals for each image.
In the basic MIL module, we followed the implementation in~\cite{tang2018pcl}, which refines the instance classifier three times.
In the SLV module, the average proposal scores of three refined instance classifiers are used to generate supervisions.
The threshold $T_{score}$ was set to 0.001 to filter out low-scoring proposals.

During training, the mini-batch size for training was set to 1.
The momentum and weight decay were set to 0.9 and $5 \times 10^{-4}$, respectively.
{The maximum iteration numbers are set to be 80K, 140K and 280K for VOC 2007, VOC 2012, and MS-COCO, respectively.
The initial learning rate was $5 \times 10^{-4}$ and the learning rate will decay at the 50K-th, 90K-th, and 220K-th iterations for VOC 2007, VOC2012, and MS-COCO, respectively.
For data augmentation, the short edges of images are randomly resized to a scale in $\left \{ 480, 576, 688, 864, 1200\right \}$, and the longest image edges are limited to 2,000.
Besides, the images are horizontally flipped.
}
During testing, the average score of 10 augmented images was used as the final classification score.

Our experiments were implemented using the PyTorch deep learning framework \cite{paszke2017automatic} and were executed on an NVIDIA PASCAL V100 GPU.

\subsection{Ablation Studies}
\begin{table}[t]
\caption{Detection performance for different ablation experiments on the PASCAL VOC 2007 $test$ set.}
\begin{center}
\begin{tabular}{c|c c c |c}
\hline
\multirow{2}{*}{Model}&SLV module&SLV module&\multirow{2}{*}{SD module}&\multirow{2}{*}{mAP$_{0.5}$}\\
                       &(Thresholding)&(Adaptive search)&&\\
\hline\hline
\textit{A}& & & & 51.1  \\
\textit{B}&\checkmark& & & 55.0  \\
\textit{C}& &\checkmark& & 55.7 \\
\textit{D}& \checkmark& &\checkmark & 56.6  \\
\textit{E}&&\checkmark&\checkmark&\bfseries{57.2} \\
\hline
\end{tabular}
\end{center}
\label{ablation-compare}
\end{table}

\begin{figure}[t]
\begin{center}
\includegraphics[width=0.8\linewidth]{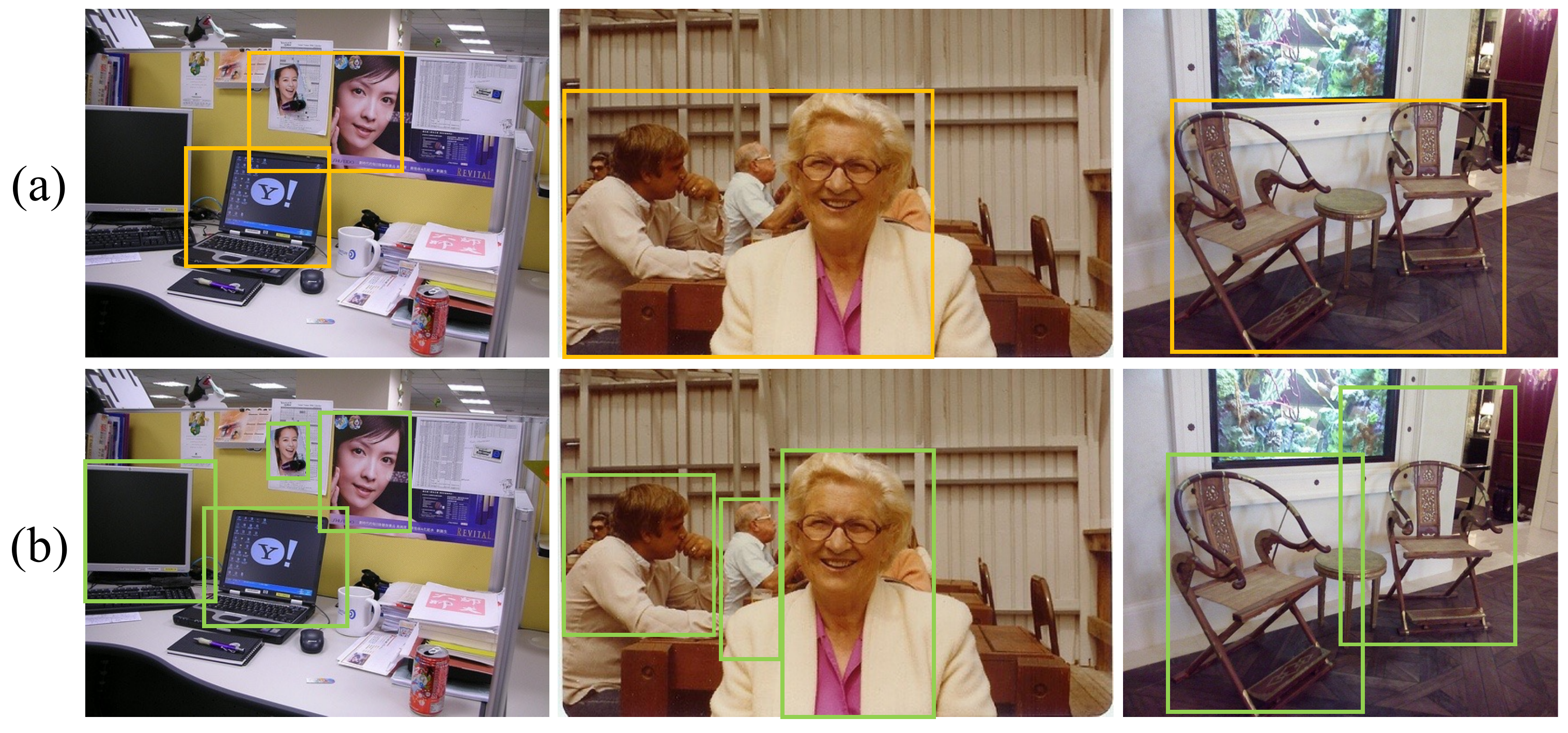}
\end{center}
\caption{Some representative labeling results of the thresholding method and adaptive search algorithm.
(a) Thresholding method.
(b) Adaptive search algorithm.}
\label{fig:search}
\end{figure}

To analyze the proposed SLV-SD Net, we conducted ablation experiments on the PASCAL VOC 2007 dataset.
The experimental and visualization results are presented in Table~\ref{ablation-compare}, Table~\ref{table:1b}, Table~\ref{table:1c}, Fig.~\ref{fig:search}, and Fig.~\ref{fig:featmap}.
{
We summarize the methods difference here for a better understanding of the following experimental results.
The PCL network proposed in~\cite{tang2018pcl} is our baseline (Model \textit{A} in Table~\ref{ablation-compare}), and our work was built on this model.
The \textit{``SLV Module (thresholding)''} (Model \textit{B} in Table~\ref{ablation-compare}) is the method proposed in our conference paper~\cite{chen2020slv}.
In this paper, we enhanced the original SLV Module with an adaptive search algorithm, namely SLV Net (Model \textit{C} in Table~\ref{ablation-compare}) in the following.
The Model \textit{E} in Table~\ref{ablation-compare} is SLV-SD Net.
}
Details of the ablation studies are discussed bellow.

\subsubsection{Effectiveness of the SLV module}
To show the effectiveness of the SLV module, we reproduce the PCL network proposed in~\cite{tang2018pcl} as our baseline and obtain 51.1\% mAP on PASCAL VOC 2007 $test$ set (Model \textit{A} in Table~\ref{ablation-compare}).
Building on Model \textit{A}, the SLV module, which uses the proposal scores of Model \textit{A} to generate the supervisions for re-classification and re-localization, is adopted.
The two versions of the SLV module both boost detection performance significantly (from 51.1\% to 55.0\% for Model \textit{B} and from 51.1\% to 55.7\% for Model \textit{C}).
These results indicate that regarding the WSOD problem as an instance classification problem is not sufficient and that the localization accuracy should be taken into consideration.
The SLV module enhances the localization abilities of the model and helps the model reach a higher level of performance.

\subsubsection{Thresholding vs. Adaptive Search}
In Table~\ref{ablation-compare}, we show the performance of two versions of the SLV module, where the thresholding method was proposed in the conference version of this paper~\cite{chen2020slv} and the adaptive search algorithm is proposed in this work.
In the thresholding method, a fixed threshold is used to binarize the likelihood map and then the minimum bounding rectangles of the connected regions in the generated binary map are regarded as the labeling results.
This method is concise but has the following two drawbacks:
(1) Because of the nature of an MIL network, the proposal scores of less salient objects are much lower than that of the most salient object. 
Therefore, the likelihood map is dominated by the most salient object and a fixed threshold setting often misses the less salient objects in the background.
(2) When there are multiple objects of the same category in a group, the thresholding method often groups these objects into one bounding box. 
As shown in Fig.~\ref{fig:search} (a), the monitor in the left part of the first image is missed because there is another more salient monitor in the middle of this image.
For the second image in Fig.~\ref{fig:search} (a), the objects of the same category (\textit{person}) are grouped into one bounding box.
In contrast to this method, the adaptive search algorithm is designed to address these two problems.
As shown in Fig.~\ref{fig:search} (b), this algorithm can find the less salient monitor in the first image and distinguish several humans and chairs in the second and third images.
The detection performances of Models \textit{B} and \textit{C} in Table~\ref{ablation-compare} also demonstrate the high performance of the adaptive search algorithm.

\begin{figure}[t]
\begin{center}
\includegraphics[width=0.8\linewidth]{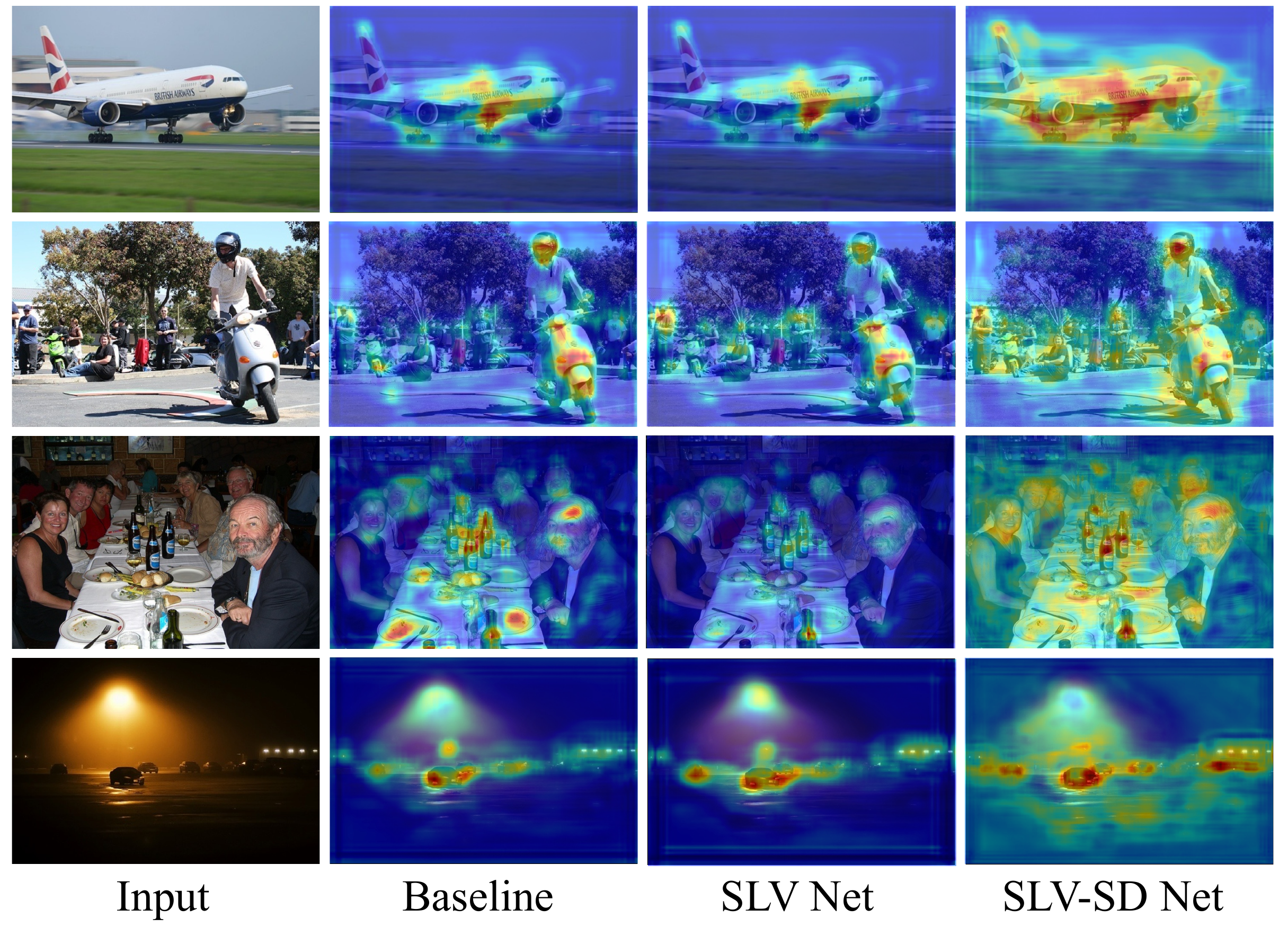}
\end{center}
\caption{{Visualization of example feature distributions of three different models.}
}
\label{fig:featmap}
\end{figure}

\subsubsection{Effectiveness of the SD module}
Under the MIL constraints, the model is prone to becoming stuck in the local minima, causing it to select the most discriminative parts of an object and ignore the full object's extent.
The SD module is proposed to encourage the model to pay more explicit attention to a wider area.
In the first row of Fig.~\ref{fig:featmap}, we show the distributions of features extracted by three different models for an image containing a single object.
For the baseline model and SLV Net, it is obvious that only a small region of the plane is activated.
Moreover, this small region is expected to play an important role in the final classification.
However, when we adopt the SD module in the SLV Net, the activated region is substantially expanded over the full extent of the plane, which implies the SLV-SD Net is able to attend to a wider area.
{In the second and third rows of Fig.~\ref{fig:featmap}, we show the feature distributions of an image containing multiple objects with different scales.
In contrast to the baseline model, SLV-SD Net activates almost the full extent of the regions of the persons, bicycles, and objects on the table, showing the significant effectiveness of the SD module.
Moreover, we select an image of the dark scene and show the features extracted by three models in the last row of Fig.~\ref{fig:featmap}.
The SLV-SD still activates more instances than other two models.
}
Based on the SLV Net, the SD module further improves detection performance with negligible computational cost, as shown in Table~\ref{ablation-compare} (from 55.7\% to 57.2\% for Model \textit{E}).

\begin{table}[t]
\begin{floatrow}
\capbtabbox{
 \begin{tabular}{c|c}
 \hline
$T_{score}$ &mAP$_{0.5}$(\%)\\
\hline\hline
0.001   &\bfseries{57.2} \\
0.005   &56.9 \\
0.01    &\bfseries{57.2} \\
0.02    &57.0 \\
0.05    &56.6 \\
0.1     &56.1 \\
\hline
 \end{tabular}
}{
 \caption{Varying $T_{score}$ ($s=5, \varepsilon=0.05$)}
 \label{table:1b}
}
\capbtabbox{
 \begin{tabular}{cc|c}
\hline
$s$ & $\varepsilon$ &mAP$_{0.5}$(\%)\\
\hline\hline
5  &0.02  &56.6 \\
10 &0.02  &56.9 \\
5  &0.05  &\bfseries{57.2} \\
10 &0.05  &57.0 \\
5  &0.1   &56.5 \\
10 &0.1   &56.3 \\
\hline
 \end{tabular}
}{
 \caption{Varying $s$ and $\varepsilon$ ($T_{score} = 0.001$)}
 \label{table:1c}
}
\end{floatrow}
\end{table}

\subsubsection{{Hyperparameter sensitivity analysis}}
Experiments were conducted to analyze the parameter sensitivity of our method.
The experimental results are shown in Table~\ref{table:1b} and Table~\ref{table:1c}.

We tried to use different $T_{score}$ and found that the performance of our method is quite stable when $T_{score}$ vary from $0.001$ to $0.02$.
However, the performance started to decline when $T_{score} > 0.02$.
This result means that those proposals with scores between $0.001$ and $0.02$ also have little significance to the final performance.
And our method is robust to the different settings of $T_{score}$.

The parameters $s$ and $\varepsilon$ were used in Algorithm~\ref{alg:algo1}.
$s$ is the search step and $\varepsilon$ determines how fast the algorithm stops.
We tried different combinations of $s$ and $\varepsilon$, and found that our method was not sensitive to $s$.
However, the parameter $\varepsilon$ cannot be set too large or too small.
Therefore, we set $\varepsilon = 0.05$ for obtaining better performance.

\begin{table*}[htb]
\caption{Average precision (in $\%$) on the PASCAL VOC 2007 test set.
The first part shows the results of weakly supervised object detectors using a single model and the second part shows the results of weakly supervised object detectors using an ensemble model or fully supervised object detector trained by pseudo ground truth generated by weakly supervised object detectors.
}
\begin{center}
\resizebox{\textwidth}{!}{
\begin{tabular}{l|cccccccccccccccccccc|c}
\hline
Method &aero&bike&bird&boat&bottle&bus&car&cat&chair&cow&table&dog&horse&mbike&person&plant&sheep&sofa&train&tv&mAP$_{0.5}$ \\
\hline\hline
PCL(VGG)~\cite{tang2018pcl}          &54.4&69.0&39.3&19.2&15.7&62.9&64.4&30.0&25.1&52.5&44.4&19.6&39.3&67.7&17.8&22.9&46.6&57.5&58.6&63.0&43.5 \\
WS-RPN(VGG)~\cite{tang2018weakly}    &57.9&70.5&37.8&5.7&21.0&66.1&69.2&59.4&3.4&57.1&\bfseries{57.3}&35.2&64.2&68.6&\bfseries{32.8}&\bfseries{28.6}&50.8&49.5&41.1&30.0&45.3 \\
C-MIL~\cite{wan2019c}                 &62.5&58.4&49.5&32.1&19.8&70.5&66.1&63.4&20.0&60.5&52.9&53.5&57.4&68.9&8.4&24.6&51.8&58.7&66.7&63.5&50.5 \\
UI~\cite{gao2019utilizing}            &63.4&70.5&45.1&28.3&18.4&69.8&65.8&69.6&27.2&62.6&44.0&59.6&56.2&\bfseries{71.4}&11.9&26.2&\bfseries{56.6}&59.6&69.2&\bfseries{65.4}&52.0 \\
Pred Net(VGG)~\cite{arun2019dissimilarity} &\bfseries{66.7}&69.5&52.8&31.4&{24.7}&{74.5}&\bfseries{74.1}&67.3&14.6&53.0&46.1&52.9&\bfseries{69.9}&70.8&18.5&28.4&54.6&60.7&67.1&60.4&52.9 \\
IM-CFB(VGG)~\cite{yin2021instance}   &64.1&74.6&44.7&29.4&26.9&73.3&72.0&71.2&\bfseries{28.1}&\bfseries{66.7}&48.1&63.8&55.5&68.3&17.8&27.7&54.4&62.7&\bfseries{70.5}&66.6&54.3 \\
MIST(+Reg.)~\cite{ren2020instance}   &-&-&-&-&-&-&-&-&-&-&-&-&-&-&-&-&-&-&-&-&54.9 \\
SLV Net(VGG)                              &62.3&\bfseries{75.0}&\bfseries{53.1}&\bfseries{35.0}&\bfseries{34.6}&\bfseries{75.4}&73.3&\bfseries{72.5}&26.5&65.5&48.0&\bfseries{65.3}&60.4&67.0&31.4&22.8&56.0&\bfseries{65.1}&70.0&55.3&\bfseries{55.7}\\
\hline\hline
PCL+FRCNN~\cite{tang2018pcl}       &63.2&69.9&47.9&22.6&27.3&71.0&69.1&49.6&12.0&60.1&51.5&37.3&63.3&63.9&15.8&23.6&48.8&55.3&61.2&62.1&48.8 \\
WS-RPN+FRCNN~\cite{tang2018weakly} &63.0&69.7&40.8&11.6&27.7&70.5&74.1&58.5&10.0&66.7&\bfseries{60.6}&34.7&\bfseries{75.7}&70.3&25.7&26.5&55.4&56.4&55.5&54.9&50.4 \\
W2F~\cite{zhang2018w2f}            &63.5&70.1&50.5&31.9&14.4&72.0&67.8&73.7&23.3&53.4&49.4&65.9&57.2&67.2&27.6&23.8&51.8&58.7&64.0&62.3&52.4 \\
UI+FRCNN~\cite{gao2019utilizing}   &62.7&69.1&43.6&31.1&20.8&69.8&68.1&72.7&23.1&65.2&46.5&64.0&67.2&66.5&10.7&23.8&55.0&62.4&69.6&60.3&52.6 \\
C-MIL+FRCNN~\cite{wan2019c}        &61.8&60.9&\bfseries{56.2}&28.9&18.9&68.2&69.6&71.4&18.5&64.3&57.2&66.9&65.9&65.7&13.8&22.9&54.1&61.9&68.2&\bfseries{66.1}&53.1 \\
Pred Net(Ens)~\cite{arun2019dissimilarity} &\bfseries{67.7}&70.4&52.9&31.3&26.1&\bfseries{75.5}&73.7&68.6&14.9&54.0&47.3&53.7&70.8&70.2&19.7&\bfseries{29.2}&54.9&61.3&67.6&61.2&53.6 \\
IM-CFB+FRCNN~\cite{yin2021instance}   &63.3&\bfseries{77.5}&48.3&\bfseries{36.0}&32.6&70.8&71.9&73.1&\bfseries{29.1}&\bfseries{68.7}&47.1&\bfseries{69.4}&56.6&\bfseries{70.9}&22.8&24.8&56.0&59.8&\bfseries{73.2}&64.6&55.8 \\
CASD~\cite{huang2020comprehensive}   &-&-&-&-&-&-&-&-&-&-&-&-&-&-&-&-&-&-&-&-&56.8 \\
\hline
SLV-SD Net(VGG)                      &63.3&74.7&53.7&35.2&\bfseries{35.6}&74.6&\bfseries{75.0}&\bfseries{77.9}&26.1&62.8&49.9&68.6&61.0&67.9&\bfseries{37.3}&27.0&\bfseries{57.3}&\bfseries{64.0}&70.7&62.3&\bfseries{57.2}\\
\hline
\end{tabular}}
\end{center}

\label{mAP-2007}
\end{table*}

\begin{table*}[htb]
\caption{CorLoc (in $\%$) on the PASCAL VOC 2007 trainval set. 
The first part shows the results of weakly supervised object detectors using a single model and the second part shows the results of weakly supervised object detectors using an ensemble model or fully supervised object detector trained by pseudo ground truth generated by weakly supervised object detectors.
}
\begin{center}
\resizebox{\textwidth}{!}{
\begin{tabular}{l|cccccccccccccccccccc|c}
\hline
Method &aero&bike&bird&boat&bottle&bus&car&cat&chair&cow&table&dog&horse&mbike&person&plant&sheep&sofa&train&tv&CorLoc \\
\hline\hline
PCL(VGG)~\cite{tang2018pcl}       &79.6&85.5&62.2&47.9&37.0&83.8&83.4&43.0&38.3&80.1&50.6&30.9&57.8&90.8&27.0&58.2&75.3&68.5&75.7&78.9&62.7 \\
WS-RPN(VGG)~\cite{tang2018weakly} &77.5&81.2&55.3&19.7&44.3&80.2&86.6&69.5&10.1&\bfseries{87.7}&\bfseries{68.4}&52.1&84.4&91.6&\bfseries{57.4}&\bfseries{63.4}&77.3&58.1&57.0&53.8&63.8 \\
C-MIL~\cite{wan2019c}              &-&-&-&-&-&-&-&-&-&-&-&-&-&-&-&-&-&-&-&-&65.0 \\
UI~\cite{gao2019utilizing}         &84.2&84.7&59.5&52.7&37.8&81.2&83.3&72.4&41.6&84.9&43.7&69.5&75.9&90.8&18.1&54.9&81.4&60.8&79.1&\bfseries{80.6}&66.9 \\
Pred Net(VGG)~\cite{arun2019dissimilarity} &\bfseries{88.6}&\bfseries{86.3}&\bfseries{71.8}&53.4&51.2&\bfseries{87.6}&\bfseries{89.0}&65.3&33.2&86.6&58.8&65.9&\bfseries{87.7}&\bfseries{93.3}&30.9&58.9&\bfseries{83.4}&67.8&78.7&80.2&70.9 \\
IM-CFB(VGG)~\cite{yin2021instance}   &-&-&-&-&-&-&-&-&-&-&-&-&-&-&-&-&-&-&-&-&70.7 \\
MIST(+Reg.)~\cite{ren2020instance}   &-&-&-&-&-&-&-&-&-&-&-&-&-&-&-&-&-&-&-&-&68.8 \\
SLV Net(VGG)                           &85.0&84.3&66.1&\bfseries{58.5}&\bfseries{58.4}&85.8&86.1&\bfseries{83.1}&\bfseries{44.9}&85.6&47.9&\bfseries{78.8}&81.3&93.2&48.1&57.5&79.4&\bfseries{69.6}&\bfseries{84.0}&73.5&\bfseries{72.6}\\
\hline\hline
PCL+FRCNN~\cite{tang2018pcl}       &83.8&85.1&65.5&43.1&50.8&83.2&85.3&59.3&28.5&82.2&57.4&50.7&85.0&92.0&27.9&54.2&72.2&65.9&77.6&\bfseries{82.1}&66.6 \\
WS-RPN+FRCNN~\cite{tang2018weakly} &83.8&82.7&60.7&35.1&53.8&82.7&88.6&67.4&22.0&86.3&\bfseries{68.8}&50.9&\bfseries{90.8}&93.6&44.0&61.2&82.5&65.9&71.1&76.7&68.4 \\
W2F~\cite{zhang2018w2f}            &-&-&-&-&-&-&-&-&-&-&-&-&-&-&-&-&-&-&-&-&70.3 \\
UI+FRCNN~\cite{gao2019utilizing}   &86.7&85.9&64.3&55.3&42.0&84.8&85.2&78.2&\bfseries{47.2}&\bfseries{88.4}&49.0&73.3&84.0&92.8&20.5&56.8&\bfseries{84.5}&62.9&82.1&78.1&66.9 \\
Pred Net(Ens)~\cite{arun2019dissimilarity} &\bfseries{89.2}&\bfseries{86.7}&\bfseries{72.2}&50.9&51.8&\bfseries{88.3}&\bfseries{89.5}&65.6&33.6&87.4&59.7&66.4&88.5&\bfseries{94.6}&30.4&60.2&83.8&\bfseries{68.9}&78.9&81.3&71.4 \\
IM-CFB+FRCNN~\cite{yin2021instance}   &-&-&-&-&-&-&-&-&-&-&-&-&-&-&-&-&-&-&-&-&72.2 \\
\hline
SLV-SD Net                   &82.5&84.7&69.1&\bfseries{59.0}&\bfseries{56.5}&86.3&87.5&\bfseries{87.2}&43.2&87.0&48.7&\bfseries{80.7}&82.3&93.6&\bfseries{56.2}&\bfseries{61.5}&78.4&66.9&\bfseries{85.6}&79.9&\bfseries{73.8}\\
\hline
\end{tabular}}
\end{center}
\label{corloc-2007}
\end{table*}

\begin{table}[t]
\caption{Comparison with other methods on the PASCAL VOC 2012 dataset.}
\begin{center}
\begin{tabular}{l|c c}
\hline
Method &mAP$_{0.5}$(\%)&CorLoc(\%)\\
\hline\hline
PCL(VGG)~\cite{tang2018pcl}          &40.6&63.2 \\
WS-RPN(VGG)~\cite{tang2018weakly}    &40.8&64.9 \\
C-MIL~\cite{wan2019c}                 &46.7&67.4 \\
UI~\cite{gao2019utilizing}            &48.0&67.4 \\
Pred Net(VGG)~\cite{arun2019dissimilarity} &48.4&69.5 \\
IM-CFB(VGG)~\cite{yin2021instance} &49.4&69.6 \\
MIST(+Reg.)~\cite{ren2020instance} &52.1&70.9 \\
CASD~\cite{huang2020comprehensive} &53.6&- \\
SLV-SD Net(VGG)                              &\bfseries{54.0}&\bfseries{72.0}\\
\hline
\end{tabular}
\end{center}
\label{results-2012}
\end{table}

\begin{table}[t]
\caption{{Comparison with other methods on the MS-COCO dataset.}}
\begin{center}
\begin{tabular}{l|c|c c}
\hline
Method &backbone&mAP&mAP$_{0.5}$\\
\hline\hline
PCL(VGG)~\cite{tang2018pcl}        &VGG16&8.5&19.4 \\
WSOD2~\cite{zeng2019wsod2}         &VGG16&10.8&22.7 \\ 
MIST(+Reg.)~\cite{ren2020instance} &VGG16&11.4&24.3 \\
MIST(+Reg.)~\cite{ren2020instance} &ResNet50&12.6&26.1 \\
CASD~\cite{huang2020comprehensive} &VGG16&12.8&26.4 \\
CASD~\cite{huang2020comprehensive} &ResNet50&\bfseries{13.9}&27.8 \\
\hline
SLV-SD Net                              &VGG16&12.4&27.3\\
SLV-SD Net                              &ResNet50&13.4&\bfseries{28.1}\\
\hline
\end{tabular}
\end{center}
\label{results-coco}
\end{table}

\subsection{Comparison with Other Methods}

In this subsection, we compare the results of the proposed SLV-SD Net with that of previous methods.
The main experimental results on the PASCAL VOC 2007 and 2012 datasets are listed in Tables~\ref{mAP-2007}, \ref{corloc-2007}, and \ref{results-2012}.
Our SLV Net with adaptive search obtains a mAP of 55.7\% with a single VGG16~\cite{simonyan2014very} model on the VOC 2007 dataset, thus outperforming all the other single-model methods.
When the SD module is adopted, the SLV-SD Net obtains a mAP of 57.2\% and produces new state-of-the-art result.
On the PASCAL VOC 2012 dataset, our SLV-SD Net obtains 54.0\% in mAP and 72.0\% in CorLoc, which is also the best of all methods.
The experimental results on the MS-COCO datasets are listed in Tables~\ref{results-coco}.
With the VGG16 backbone, SLV-SD Net obtains 12.4\% mAP and 27.3\% mAP$_{0.5}$.
Our method outperforms the VGG version of CASD~\cite{huang2020comprehensive} by 0.9\% in mAP$_{0.5}$ but is inferior to it in mAP (12.4\% \textit{vs.} 12.8\%).
Similarly, with the ResNet50~\cite{he2016deep} backbone, our method outperforms CASD in mAP$_{0.5}$, but CASD is better than our method in mAP.
CASD~\cite{huang2020comprehensive} performs better when using higher IoU thresholds.
Our method performs the best when IoU threshold is 0.5.
This result suggests that a sophisticated feature refining procedure like CASD should be a promising exploration for obtaining high-quality detector.

\begin{figure*}
\begin{center}
\includegraphics[width=1.0\linewidth]{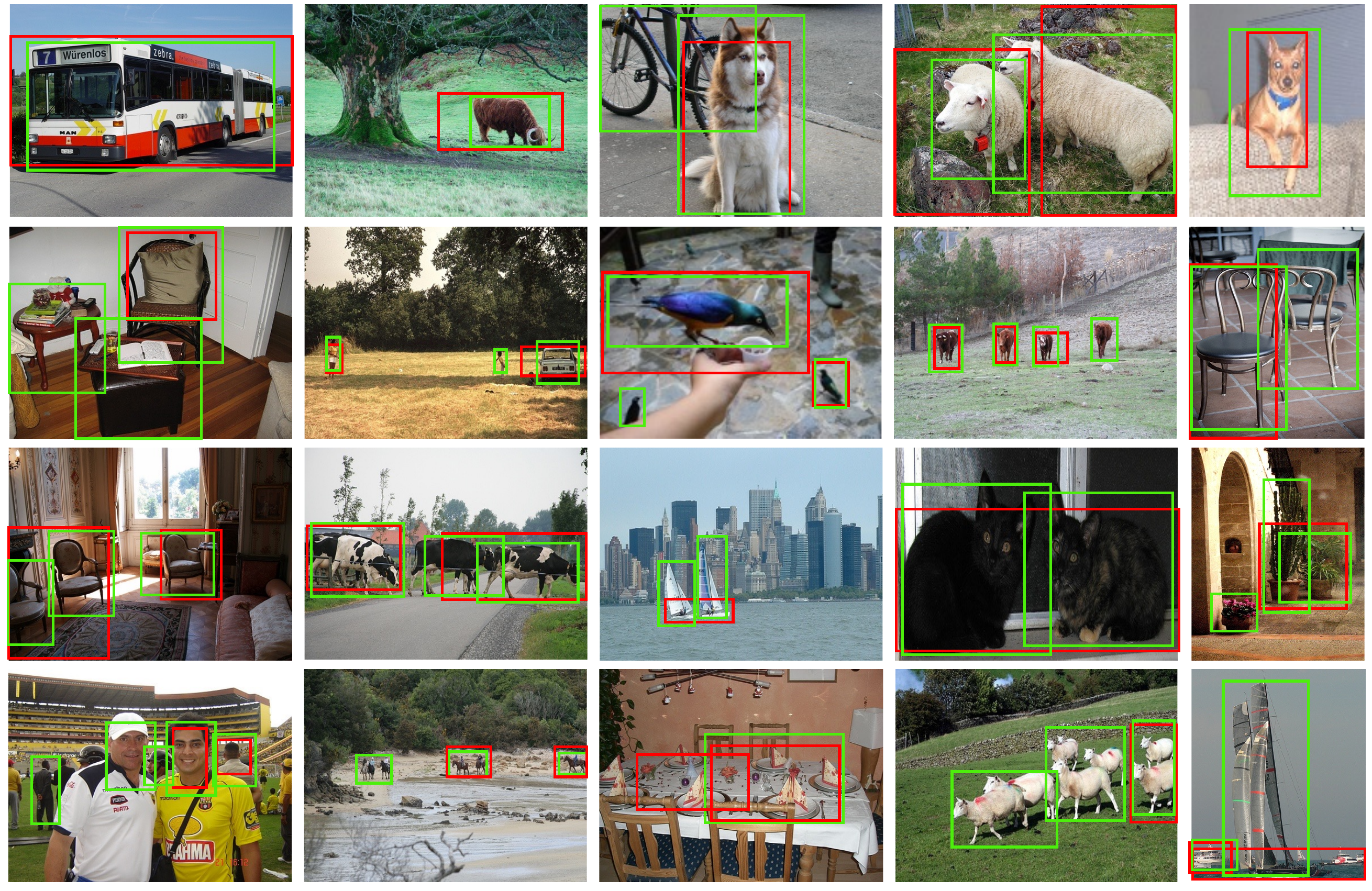}
\end{center}
   \caption{Detection results of our method and the PCL model.
   Green bounding boxes are the objects detected by our method and the red ones are the results detected by PCL.}
\label{fig:final result}
\end{figure*}

In contrast to recent methods, \textit{e.g.} \cite{yang2019towards,ren2020instance}, which select high-scoring proposals as pseudo ground-truth to enhance localization ability, SLV-SD Net is devoted to searching the boundaries of objects from a more macroscopic perspective.
The aim of SLV is to find annotations for each object while the aim of SD is to refine the feature distributions of images, which can help the former perform better.
We illustrate some typical detection results of our method and a competing model (PCL) in Fig.~\ref{fig:final result}.
In the first row of Fig.~\ref{fig:final result}, we present images containing recognizable objects.
It is obvious that the bounding boxes output by our method are better localized.
The images in the second row of Fig.~\ref{fig:final result} contain objects that are difficult to distinguish.
Our method detects almost all the instances in these images (the person’s hand in the third image is missed) and obtains improved localization performance.
The third row of Fig.~\ref{fig:final result} presents images containing objects that are close together.
While the competing model detects such objects into one bounding box, our method is able to detect them separately.
Although our method substantially outperforms the competitor, we show the failure cases of our method in the last row of Fig.~\ref{fig:final result}.
From these failure cases, we can summarize that our method cannot bring much promotion for those indistinguishable categories (\textit{e.g.} table, chair) and does not perform well when there are heavily clustered objects in an image.
Addressing these issues will be the focus of our future research.

\section{Conclusion}
This paper presented an effective WSOD framework called SLV-SD Net.
We proposed reformulating the instance classification problem in most MIL-based models into a multi-tasking problem to close the gap between weakly supervised and fully supervised object detection.
The SLV module uses region proposal scores to generate accurate instance-level annotations and thus converges the region proposal localization in the model.
Using the SLV module, the SD module refines the feature representations of images, which in turn improves detection performance.
The results of extensive experiments conducted on the PASCAL VOC 2007/2012 and MS-COCO datasets show that SLV-SD Net substantially improves performance, outperforming other state-of-the-art models.

\section*{Acknowledgements}
This work was supported by Alibaba Group through Alibaba Research Intern Program.
And this work was supported by the Fundamental Research Funds for the Central Universities, and the National Natural Science Foundation of China under Grant 31627802.



\bibliographystyle{elsarticle-num}
\bibliography{mybib}





\end{document}